\documentclass[runningheads]{llncs}
\usepackage{graphicx}
\usepackage{adjustbox}
\newcommand{\ie}{{i}.{e}.}
\newcommand{\eg}{{e}.{g}.}
\newcommand{\eal}{{\it et al.}}

\usepackage[table,xcdraw]{xcolor}
\usepackage{url}
\usepackage{multirow}
\usepackage{amsmath}
\usepackage{makecell}
\usepackage{array}
\usepackage{fixltx2e}

\usepackage{arydshln}
\usepackage{mathtools}
\usepackage{booktabs}
\makeatletter
\newcommand\figcaption{\def\@captype{figure}\caption}
\newcommand\tabcaption{\def\@captype{table}\caption}
\makeatother
\usepackage[misc]{ifsym}

\usepackage{xcolor}
\definecolor{blue}{rgb}{0,0,1}

\definecolor{red}{rgb}{1,0,0}

\definecolor{black}{rgb}{0,0,0}

\usepackage{hyperref}
\hypersetup{colorlinks=true,citecolor=blue,linkcolor=blue,linktocpage=true}

\begin{document}
\title{Deblurring Masked Autoencoder is Better Recipe for Ultrasound Image Recognition \thanks {Code will be available at: \href{https://github.com/MembrAI/DeblurringMIM}{https://github.com/MembrAI/DeblurringMIM} }}
\titlerunning{Deblurring Masked Autoencoder}
%
\author{Qingbo Kang 
\inst{1,3} 
\and Jun Gao 
\inst{1, 4} 
\and Kang Li 
\inst{1,3} \textsuperscript{\Letter}
\and Qicheng Lao 
\inst{2,3} \textsuperscript{\Letter}
}
\authorrunning{Q. Kang et al.}

\institute{ West China Biomedical Big Data Center, West China Hospital, Sichuan University \and
School of Artificial Intelligence, Beijing University of Posts and Telecommunications, Beijing, China \and 
 Shanghai Artificial Intelligence Laboratory, Shanghai, China \and
College of Computer Science, Sichuan University, Chengdu, China 
}
%
%
\maketitle              
\begin{abstract}
Masked autoencoder (MAE) has attracted unprecedented attention and achieves remarkable performance in many vision tasks. It reconstructs random masked image patches (known as proxy task) during pretraining and learns meaningful semantic representations that can be transferred to downstream tasks. However, MAE has not been thoroughly explored in ultrasound imaging. In this work, we investigate the potential of MAE for ultrasound image recognition. Motivated by the unique property of ultrasound imaging in high noise-to-signal ratio, we propose a novel deblurring MAE approach that incorporates deblurring into the proxy task during pretraining. The addition of deblurring facilitates the pretraining to better recover the subtle details presented in the ultrasound images, thus improving the performance of the downstream classification task. Our experimental results demonstrate the effectiveness of our deblurring MAE, achieving state-of-the-art performance in ultrasound image classification. Overall, our work highlights the potential of MAE for ultrasound image recognition and presents a novel approach that incorporates deblurring to further improve its effectiveness.

\end{abstract}
\section{Introduction}

Recently, as representative of generative self-supervised learning (SSL) methods, masked autoencoder (MAE) \cite{he2022masked} has achieved great success in many vision tasks~\cite{li2022exploring,ke2022mask,wang2022contrastmask}. In general, MAE belongs to the masked image modeling (MIM) paradigm \cite{zhang2022survey}, where some parts of the image are randomly masked, and the purpose of pretraining (\ie, proxy or pretext task) is to recover the missing pixels. After the pretraining, the learned image representation can be transferred to downstream tasks for improved performance. 
With the advent of MAE, many MAE variants have been proposed \cite{tian2022beyond,wu2022denoising,gao2022convmae}. Tian~\eal~\cite{tian2022beyond} investigate other image degradation methods during MAE pretraining and find that the optimal practice is enriching masking with spatial misalignment for nature images. Wu~\eal~\cite{wu2022denoising} design a denoising MAE by introducing Gaussian noising into MAE pretraining, showing that their denoising MAE is robust to additive noises. 

On the other hand, although numerous work has been proposed for applying MAE to medical imaging across different modalities including pathological images \cite{quan2022global,luo2022self,an2022masked}, X-rays \cite{zhou2022self,xiao2023delving}, electrocardiogram \cite{zhang2022maefe}, immunofluorescence images \cite{ly2022student}, MRI and CT \cite{zhou2022self,xu2022swin,chen2023masked}. However, the majority of them have not fully exploited the characteristics of medical images and instead, focus on vanilla applications~\cite{zhou2022self,zhang2022maefe,xiao2023delving,chen2023masked}. 
This is especially problematic given the domain gap between medical and natural images, as well as the unique imaging properties associated with each medical imaging modality \cite{raghu2019transfusion,niu2021distant,qin2022medical}. Furthermore, as an important and widely used medical imaging modality, ultrasound has not been extensively explored in the context of MAE-based approaches.

Based on the aforementioned analysis, in this paper, we propose a deblurring masked auto-encoder framework, which is specifically designed for ultrasound image recognition. The primary motivation for the deblurring comes from the unique imaging properties of ultrasound, \eg, high noise-to-signal ratio. Compared with nature images, the subtle details within ultrasound are particularly important for downstream analysis (\eg, microcalcifications is an important sign for malignant nodules, which is represented as tiny bright spots in ultrasound \cite{taki2004thyroid,park2009sonography}). Moreover, the motivation also stems from the findings of our preliminary experiments, 
which suggest that denoising may not be appropriate for inherently noisy ultrasound images. Therefore, we introduce the opposite direction with a deblurring approach for ultrasound images. Specifically, we first apply blurring operations to the ultrasound images prior to the random masking during pretraining, enabling the model to learn how to de-blur and reconstruct the original image. It should be emphasized that denoising and deblurring are two opposite directions, \ie, denoising first adds noise to the clean image and learns to remove the noise, while deblurring blurs the noisy ultrasound image and learns to sharpen the image. The deblurring facilitates the pretraining in recovering the subtle details within the image, which is crucial for ultrasound image recognition. It should be emphasized that while blurring operation has been shown ineffective for natural images \cite{tian2022beyond}, ultrasound images are fundamentally different and may benefit from blurring operation.

Furthermore, to the best of our knowledge, this paper is the first attempt to apply the MAE approach to ultrasound image recognition. Our work also addresses some fundamental concerns that are of great interest to the medical imaging community with the example of ultrasound, such as the importance of in-domain data pretraining for MAE in ultrasound, as well as the finding that SSL pretraining is consistently better than the supervised pretraining as with nature images. To conclude, our contributions can be summarized as follows:\begin{enumerate}
    \item 
    We propose a deblurring MAE framework that is specifically designed for ultrasound images by incorporating a deblurring task into MAE pretraining. This is motivated by the fact that ultrasound images have a high noise-to-signal ratio, and in contrast to denoising for natural images, we demonstrate that deblurring is a better recipe for ultrasound images. 
    \item We explore the effectiveness of various image blurring methods in our deblurring MAE and find that a simple Gaussian blurring performs the best, showing superior transferability compared with the vanilla MAE.
    \item We conduct experiments on more than 10k ultrasound images for pretraining and 4,494 images for downstream thyroid nodule classification. The results demonstrate the effectiveness of the proposed deblurring MAE, achieving state-of-the-art classification performance for ultrasound images.

\end{enumerate}

Note that, as a representative MIM approach, the MAE is adopted to validate our proposed deblurring pretraining in this work, our method can also be seamlessly integrated with other MIM-based approaches such as ConvMAE \cite{gao2022convmae}. 

\section{Method}
\subsection{Preliminary: MAE}
The MAE pipeline consists of two primary stages: self-supervised pretraining and transferring for downstream tasks. During the self-supervised pretraining, the model is trained to reconstruct masked input image patches using an asymmetric encoder-decoder architecture. The encoder is typically a ViT \cite{dosovitskiy2020image}, which compresses the input image into a latent representation, while the decoder is a lightweight Transformer that reconstructs the original image from the latent representation. The loss used during pretraining is the mean squared error (MSE) between the reconstructed and original images. In the transfer stage, the weights of the pre-trained ViT encoder are transferred and used as a feature extractor, to which task-specific heads are appended for learning various downstream tasks. Typically, there are two common practices in the transfer stage: 1) end-to-end fine-tuning which tunes the entire model, and 2) linear probing, which only tunes the task-specific head.

\subsection{Our Proposed Deblurring MAE}
Similar to MAE, our proposed deblurring MAE also contains pretraining and transfer learning for downstream tasks. We employ the same asymmetric encoder-decoder architecture as the original MAE.

\subsubsection{Deblurring MAE Pretraining}
\begin{figure}[t!]
  \centering
  \includegraphics[width=\textwidth]{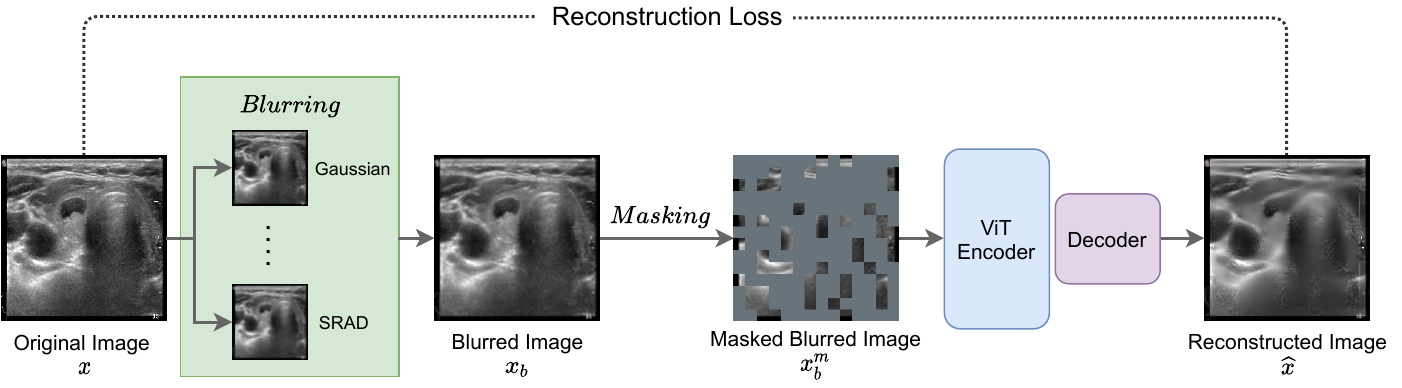}
  \caption{Illustration of our proposed deblurring MAE pretraining.}
  \vspace{-12pt}
  \label{fig_main}
\end{figure}

For the pretraining, besides the original masked image modeling task in the MAE, we introduce one additional task, \ie, deblurring, into the pretraining thus making the pretraining as deblurring pretraining. As shown in Figure~\ref{fig_main}, the deblurring is achieved by simply inserting an image blurring operation prior to random masking. 
The pipeline of our deblurring MAE pretraining is illustrated in Eq.~\ref{eq_1}:
\begin{equation}
\label{eq_1}
x \xrightarrow[]{Blurring} x_b \xrightarrow[]{Masking} x_b^m \xrightarrow[]{\text{ViT Encoder}} h \xrightarrow[]{\text{Decoder}} \hat{x} .
\end{equation}
Specifically, the original ultrasound image $x$ is first blurred by a chosen image blurring operation $Blurring$ to obtain $x_b$. After that, several patches in the blurred image $x_b$ are randomly masked by the $Masking$ operation with a pre-defined ratio to obtain $x_b^m$. Next, the masked blurred image $x_b^m$ is passed as input to the $\text{ViT Encoder}$, which generates a latent representation $h$. Finally, the $\text{Decoder}$ receives the representation $h$ and outputs reconstructed image $\hat{x}$. 

The image blurring operation $Blurring$ is a commonly used technique for reducing the sharpness or details of an image, resulting in a smoother, less-detailed appearance. There exist many different methods for image blurring, with most of them involving the averaging of neighboring pixels in some way. In Figure~\ref{fig_main}, we provide examples of two representative blurring methods: Gaussian blur and speckle reducing anisotropic diffusion (SRAD) \cite{yu2002speckle}. 

Gaussian blur involves convolving an input image with a Gaussian kernel $G(\sigma)$, which is a two-dimensional Gauss function that represents a normal distribution with standard deviation of $\sigma$. Mathematically, Gaussian blur can be defined as follows:
\begin{equation}
    x_b = Gaussian(x, \sigma) = x \ast G(\sigma) = x \ast \frac{1}{2\pi \sigma^2} e^{-(u^2+v^2)/2\sigma^2} ,
\end{equation}
where $\ast$ denotes the convolution operation, and $(u, v)$ represents the coordinates in the kernel. The degree of blurring (\ie, blurriness) in the resulting image is determined by the standard deviation $\sigma$.

The SRAD is a nonlinear anisotropic diffusion technique for removing speckled noises, which has been extensively used in medical ultrasound images, due to its edge-sensitivity for speckled images and powerful preservation of useful information. The SRAD operation is implemented by repeating an anisotropic diffusion equation for $N$ iterations. It can be formally given as:
\begin{equation}
\label{eq_srad_1}
    x_b = SRAD(x, N, t) = x(i, j, 0) + \Delta t * \sum_{k=0}^{N-1} \text{div}(\text{c}(i, j, k) \nabla x (i, j, k) )
,
\end{equation}
where $x$ is the original image, $N$ stands for the number of iterations, $t$ means time. $x(i, j, k)$ and $\text{c}(i, j, k)$ represent the image and diffusivity coefficient at iteration $k$, respectively. $\nabla x$ is the gradient of $x$ and $\text{div}$ is the divergence operator. The larger $N$ or $t$ leads to a blurrier resulting image.


The pixel-wise MSE between the reconstructed image $\hat{x}$ and the original image $x$ is utilized as the loss function during pretraining: $\mathcal{L}_{MSE} = ||\hat{x} - x||_{2}$. It should be noted that a key difference from MAE is that we compute the loss across all patches, including the masked ones. This operation is necessary due to the fact that our blurring operation covers the entire image. Through the use of the proposed deblurring MAE pretraining, we aim to leverage both masked image modeling and deblurring in order to learn a robust and effective latent representation that could be successfully applied to a range of downstream tasks.

\vspace{-8pt}
\subsubsection{Deblurring MAE Transfer}
After the deblurring MAE pretraining, only the pre-trained encoder is transferred to the downstream thyroid nodule classification task. One multi-layer perceptron (MLP) head is appended after the pre-trained encoder. The transfer learning pipeline is shown in Eq.~\ref{eq_transfer}:
\begin{equation}
\label{eq_transfer}
    x \xrightarrow[]{Blurring} x_b \xrightarrow[]{\text{ViT Encoder}} h \xrightarrow[]{\text{MLP}} \hat{y} ,
\end{equation}
It should be noted here that, in order to prevent data distribution shift between pretraining and transfer stages, the original image $x$ also needs to be blurred before fed into the pre-trained encoder during transfer learning. The cross-entropy loss  between ground-truth classification label $y$ and predicted label $\hat{y}$ is used as the loss function: $\mathcal{L}_{CE} = -[y\log(\hat{y}) + (1-y)\log(1-\hat{y})]$.

\section{Experiments and Results}

\subsection{Experimental Settings}
\subsubsection{Dataset} 
All thyroid ultrasound images used in our study for both pretraining and downstream classification were acquired at West China Hospital with ethical approval. We use a total of 10,675 images for pretraining and 4,493 images for the downstream classification. To avoid any potential data leakage, the images used in pretraining were not included in the test set for thyroid nodule classification.  
The downstream classification dataset contains 2,576 benign and 1,917 malignant cases. We randomly split the dataset into train/validation/test subsets with a 3:1:1 ratio. The classification ground-truth labels were obtained either from the fine-needle aspiration for malignant nodules or clinical diagnosis by senior radiologists for benign nodules.

\vspace{-10pt}
\subsubsection{Implementation Details}
We use a mask ratio of 75\% during the pretraining. We set the batch size to 256 for both pretraining and end-to-end fine-tuning, and 1024 for linear probing. The epochs of pretraining is 12,000 due to our relatively small data. The full detailed experimental settings are presented in the appendix. We implement our approach based on PyTorch. 
The image size for both pretraining and transfer learning is $224\times224$. For classification, we choose the model that performs the best on the validation set as the final model to evaluate on the test set. Three widely used metrics accuracy (ACC), F1-score (F1), and the area under the receiver operating characteristic (AUROC) are utilized for classification performance evaluation.

\subsection{Results and Comparisons}

\subsubsection{Our Deblurring MAE vs. vanilla MAE}
\begin{figure}[t!] 
\begin{minipage}[b]{0.62\linewidth} 
    \centering 
    \includegraphics[width=\linewidth]{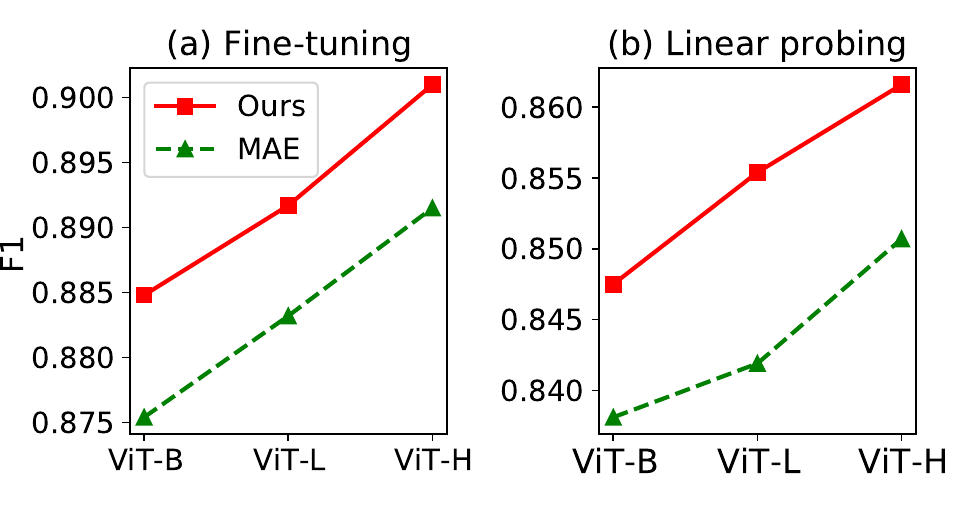} 
    \vspace{-6ex}
    \figcaption{Our deblurring MAE vs. vanilla MAE. Pre-trained with the same ultrasound data. } 
    \label{fig_compare} 
  \end{minipage}%
\begin{minipage}[b]{0.38\linewidth} 
\centering
\begin{tabular}{|lccc|}
\hline
\multicolumn{2}{|c|}{Image Blurring} &
\multicolumn{2}{|c|}{Blurriness}
 \\ \hline
\multicolumn{1}{|c|}{Method}       &  \multicolumn{1}{|c|}{F1 (\%)}                              & \multicolumn{1}{c|}{$\sigma$ }    &  \multicolumn{1}{|c|}{F1 (\%)}             \\ \hline
\multicolumn{1}{|c|}{Gaussian} & \multicolumn{1}{c|}{\textbf{88.48}} & \multicolumn{1}{c|}{0.8}  & 88.03          \\
\multicolumn{1}{|c|}{SRAD}     & \multicolumn{1}{c|}{88.13}          & \multicolumn{1}{c|}{1.1}  & \textbf{88.48} \\
\multicolumn{1}{|c|}{Mean}     & \multicolumn{1}{c|}{88.11}          & \multicolumn{1}{c|}{1.4}  & 87.56          \\
\multicolumn{1}{|c|}{Median}   & \multicolumn{1}{c|}{87.63}          & \multicolumn{1}{c|}{1.7}  & 85.89                                  \\
\multicolumn{1}{|c|}{Motion}   & \multicolumn{1}{c|}{78.33}                                  & \multicolumn{1}{c|}{2.0} & 84.74                                  \\
\multicolumn{1}{|c|}{Defocus}  & \multicolumn{1}{c|}{85.37}                                  & \multicolumn{1}{c|}{2.3} & 84.62                                  \\ \hline \hline
\multicolumn{1}{|c|}{Baseline}     & \multicolumn{3}{c|}{87.54}              \\ \hline
\end{tabular}%

\tabcaption{Ablation study.}
\label{table_comp}
\end{minipage} 
\end{figure}

First of all, in order to evaluate the effectiveness of the proposed deblurring MAE for ultrasound images, we compare the transfer learning performance between our deblurring MAE and the vanilla MAE. Table~\ref{table_main} and Figure~\ref{fig_compare} give the classification performance comparison of these two approaches. For our deblurring MAE, we use Gaussian blurring with $\sigma$ equal to 1.1 as the blurring operation. In Figure~\ref{fig_compare}, we report the experimental results of three models: ViT-Base (ViT-B), ViT-Large (ViT-L) and ViT-Huge (ViT-H), and two transfer learning paradigms: end-to-end fine-tuning and linear probing. As shown in the figure, both the fine-tuning and linear probing performance of our proposed deblurring MAE is consistently better than that of the vanilla MAE, which indicates the effectiveness of deblurring for enhancing the transferability of learned representations during ultrasound pretraining.

\vspace{-8pt}
\subsubsection{Comparison with state-of-the-art approaches}
\begin{table}[t!]
\centering
\caption{Performance comparison of different methods.}
\vspace{-6pt}
\label{table_main}
\resizebox{0.98\textwidth}{!}{%
\begin{tabular}{|c|l|l|c|c|c|c|}
\hline
\multicolumn{1}{|l|}{{\color[HTML]{000000} }}                                                                    &  \multicolumn{1} {c|} {   {\color[HTML]{000000} Method}}      & \multicolumn{1}{c|}{{\color[HTML]{000000} Architecture}} & {\color[HTML]{000000} Pretraining} & {\color[HTML]{000000} ACC (\%)}     & {\color[HTML]{000000} F1 (\%)} & {\color[HTML]{000000} AUROC (\%)} \\ \hline
{\color[HTML]{000000} }                                                                                        & {\color[HTML]{000000} ResNet \cite{he2016deep}}                           & {\color[HTML]{000000} ResNet-101}                        & {\color[HTML]{000000} -}                   & {\color[HTML]{000000} 86.06$\pm$0.87}  & {\color[HTML]{000000} 83.18$\pm$}1.15 & {\color[HTML]{000000} 91.96$\pm$1.47}  \\ \cline{2-7} 
{\color[HTML]{000000} }                                                                                        & {\color[HTML]{000000} ConvNeXt \cite{liu2022convnet}}                         & {\color[HTML]{000000} ConvNeXt-L}                        & {\color[HTML]{000000} ImageNet}            & {\color[HTML]{000000} 87.76$\pm$0.66}  & {\color[HTML]{000000} 85.47$\pm$0.91} &{\color[HTML]{000000} 93.22$\pm$1.10}  \\ \cline{2-7} 
{\color[HTML]{000000} }                                                                                        & {\color[HTML]{000000} Swin Transformer \cite{liu2021swin}}                 & {\color[HTML]{000000} Swin-L}                            & {\color[HTML]{000000} ImageNet}            & {\color[HTML]{000000} 87.43$\pm$0.68}  & {\color[HTML]{000000} 84.92$\pm$0.82} &{\color[HTML]{000000} 92.83$\pm$1.02}  \\ \cline{2-7} 
{\color[HTML]{000000} }                                                                                        & {\color[HTML]{000000} Wang \eal \cite{wang2018simultaneous}}                             & \multicolumn{1}{c|}{{\color[HTML]{000000} -}}            & {\color[HTML]{000000} -}                   & {\color[HTML]{000000} 87.44$\pm$0.75}  & {\color[HTML]{000000} 85.16$\pm$0.87} & {\color[HTML]{000000} 93.11$\pm$1.09}  \\ \cline{2-7} 
{\color[HTML]{000000} }                                                                                        & {\color[HTML]{000000} Zhou \eal \cite{zhou2021multi}}                             & \multicolumn{1}{c|}{{\color[HTML]{000000} -}}            & {\color[HTML]{000000} -}                   & {\color[HTML]{000000} 88.15$\pm$0.67}  & {\color[HTML]{000000} 86.09$\pm$0.74} & {\color[HTML]{000000} 94.17$\pm$1.21}  \\ \cline{2-7} 
{\color[HTML]{000000} }                                                                                        & {\color[HTML]{000000} }                                 & {\color[HTML]{000000} }                                  & {\color[HTML]{000000} -}                   & {\color[HTML]{000000} 80.60$\pm$}1.62  & {\color[HTML]{000000} 76.98$\pm$2.05} & {\color[HTML]{000000} 83.89$\pm$2.89}  \\
\multirow{-7}{*}{{\color[HTML]{000000} Supervised}}                                                            & \multirow{-2}{*}{{\color[HTML]{000000} ViT \cite{dosovitskiy2020image}}}            & \multirow{-2}{*}{{\color[HTML]{000000} ViT-B}}        & {\color[HTML]{000000} ImageNet}            & {\color[HTML]{000000} 86.38$\pm$0.74}  & {\color[HTML]{000000} 84.17$\pm$0.98} & {\color[HTML]{000000} 92.69$\pm$0.48}  \\ \hline
{\color[HTML]{000000} }       & {\color[HTML]{000000} SimCLR \cite{chen2020simple} }                    & {\color[HTML]{000000} ResNet-50 }                                  & {\color[HTML]{000000} ImageNet}          & {\color[HTML]{000000} 86.21$\pm$0.96}  & {\color[HTML]{000000} 83.81$\pm$1.24} & {\color[HTML]{000000} 92.16$\pm$1.08}  \\ \cline{2-7}  
{\color[HTML]{000000} }                                                                                   & {\color[HTML]{000000} }                                 & {\color[HTML]{000000} }                                  & {\color[HTML]{000000} ImageNet}            & {\color[HTML]{000000} 86.96$\pm$0.85} & {\color[HTML]{000000} 84.48$\pm$1.12} & {\color[HTML]{000000} 92.77$\pm$0.67}  \\
{\color[HTML]{000000} }                                                                                        & \multirow{-2}{*}{{\color[HTML]{000000} MoCo v3 \cite{chen2021empirical}}}        & {\color[HTML]{000000} }                                  & {\color[HTML]{000000} Ultrasound}          & {\color[HTML]{000000} 87.08$\pm$0.78} & {\color[HTML]{000000} 84.55$\pm$1.04} & {\color[HTML]{000000} 92.95$\pm$0.59}  \\ \cline{2-2} \cline{4-7} 
{\color[HTML]{000000} }                                                                                        & {\color[HTML]{000000} }                                 & {\color[HTML]{000000} }                                  & {\color[HTML]{000000} ImageNet}            & {\color[HTML]{000000} 87.25$\pm$0.51}  & {\color[HTML]{000000} 85.23$\pm$0.57}  & {\color[HTML]{000000} 93.71$\pm$0.60} \\
{\color[HTML]{000000} }                                                                                        & \multirow{-2}{*}{{\color[HTML]{000000} MAE \cite{he2022masked}}}            & {\color[HTML]{000000} }                                  & {\color[HTML]{000000} Ultrasound}          & {\color[HTML]{000000} 89.45$\pm$0.53}  & {\color[HTML]{000000} 87.54$\pm$0.62}   & {\color[HTML]{000000} 95.54$\pm$0.46}\\ \cline{2-2} \cline{4-7} 
{\color[HTML]{000000} }                                                                                        & {\color[HTML]{000000} Denoising MAE}                    & {\color[HTML]{000000} }                                  & {\color[HTML]{000000} Ultrasound}          & {\color[HTML]{000000} 80.38$\pm$1.37}  & {\color[HTML]{000000} 77.99$\pm$1.75} & {\color[HTML]{000000} 84.38$\pm$2.13}  \\ \cline{2-2} \cline{4-7} 
{\color[HTML]{000000} }                                                                                        & {\color[HTML]{000000} Ours {[}SRAD{]} } & {\color[HTML]{000000} }                                  & {\color[HTML]{000000} Ultrasound}          & {\color[HTML]{000000} 90.07$\pm$0.47}  & {\color[HTML]{000000} 88.13$\pm$0.51} & {\color[HTML]{000000} 95.87$\pm$0.45}   \\
\multirow{-7}{*}{{\color[HTML]{000000} \begin{tabular}[c]{@{}c@{}}SSL \end{tabular}}} & Ours {[}Gaussian{]}      & \multirow{-6}{*}{{\color[HTML]{000000} ViT-B}}        & Ultrasound                                 & \textbf{90.19$\pm$0.47}                & \textbf{88.48$\pm$0.50} & {\color[HTML]{000000} \textbf{96.08$\pm$0.41}}         \\   \hline    
\end{tabular}%
}
\vspace{-10pt}
\end{table}

Secondly, we also compare our approach with more approaches and the results are listed in Table~\ref{table_main}. We implement two variants of our deblurring MAE which differ in blurring operation: the SRAD with $N$ equals to 40 and $t$ equals to 0.1, and the Gaussian blur with $\sigma$ equals to 1.1. We compare with methods based on supervised learning or self-supervised learning. 
In addition, we still add the denoising MAE for comparison, although it has proved to be ineffective for ultrasound images based on our preliminary experiments. We adopt ViT-B as the architecture for these SSL-based methods except SimCLR \cite{chen2020simple} which uses ResNet-50, and we use two types of data for pretraining, \ie, ImageNet \cite{deng2009imagenet} and ultrasound. The results are based on end-to-end fine-tuning. According to Table~\ref{table_main}, we can draw the following conclusions: 

\noindent \textbf{The deblurring MAE pretraining can improve the transferability of learned representations.} First of all, both the two variants of our proposed approach (Ours [SRAD] and Ours [Gaussian]) obtain much higher classification metrics compared with the MAE pretrained using ultrasound, which indicates the learned representation of our deblurring MAE is more effective than the vanilla MAE when transferred to downstream classification. 
In addition, Table~\ref{table_main} also shows that the performance of our proposed deblurring MAE with Gaussian blurring achieves state-of-the-art performance in terms of all metrics, surpassing all competing SSL or supervised-based approaches, which further demonstrates the superior performance of our proposed deblurring MAE. It is noteworthy that, as the opposite approach to our deblurring MAE, the denoising MAE obtains worse performance compared with the vanilla MAE, suggesting that adding noise to ultrasound images during MAE pretraining is unfavorable.

\begin{figure}[t!]
  \centering
  \includegraphics[width=\textwidth]{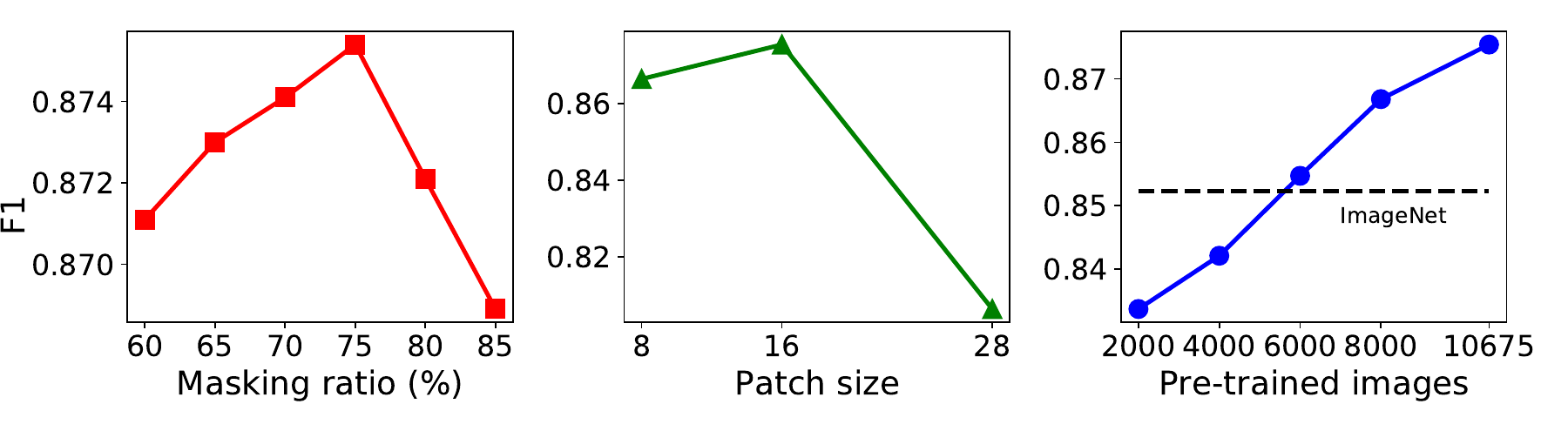}
  \vspace{-20pt}
  \caption{Hyper-parameter choices for MAE pretraining.}
  \label{fig_hyper}
  \vspace{-10pt}
\end{figure}

\noindent \textbf{Ultrasound pretraining is better than ImageNet pretraining, better than supervised pretraining.} Table~\ref{table_main} shows that the performance of MAE with ultrasound pretraining is better than the ImageNet pretraining, which underlines the importance of in-domain self-supervised pretraining in MAE. In contrast to MAE, our experiments show that the MoCo v3~\cite{chen2021empirical} achieves only marginal improvement with ultrasound pretraining. Furthermore, the MAE ImageNet pretraining also performs much better than the ImageNet supervised pretraining. These two conclusions are consistent with other works \cite{xiao2023delving,he2022masked}.

\vspace{-8pt}
\subsubsection{Ablation Study}
We design two sets of ablation studies, \ie, different image blurring methods, and the degree of blurring (blurriness) used in our deblurring MAE. We adopt the ViT-B as the architecture and end-to-end fine-tuning in transfer learning for the ablation experiments. Table~\ref{table_comp} presents the performance results of the ablation study, where the `Baseline' represents the vanilla MAE.

Firstly, besides the Gaussian and SRAD, we also try several other blurring methods that are commonly used in the fields including mean, median, motion and defocus blur. We set the kernel size to 5 in mean, median and motion blur, and the radius of defocusing is set to 5 for defocus blur. The performance results are presented on the left side of Table~\ref{table_comp}. From this table, we can observe that the Gaussian blurring achieves the best F1. And these six blurring methods are not all beneficial for pretraining, where some of them (motion, defocus) perform even worse than the baseline. 
Secondly, to investigate the effect of blurriness on the pretraining, we conduct ablation experiments on blurriness based on Gaussian blurring. 
The right side of Table~\ref{table_comp} reports the performance results and we can see that the $\sigma$ with 1.1 obtains the highest F1. In addition, as the $\sigma$, \ie, the blurriness continues to increase, the performance drops rapidly, which indicates that only a limited range of blurriness has a positive effect on the pretraining.

\vspace{-8pt}
\subsubsection{Hyper-parameter choices for MAE pretraining}

We conduct experiments to explore hyper-parameter choices for MAE pretraining based on ViT-B, and the results are presented in Figure~\ref{fig_hyper}. Our findings indicate that a masking ratio of 75\% and a patch size of 16 achieve the best transfer performance, which is consistent with MAE for natural images \cite{he2022masked}. Additionally, we observed that transfer performance improves with an increase in pre-trained images, surpassing ImageNet transfer only when a substantial amount of pre-trained images is used.

\vspace{-8pt}
\subsubsection{Visualization}
\begin{figure}[t!]
  \centering
  \includegraphics[width=\textwidth]{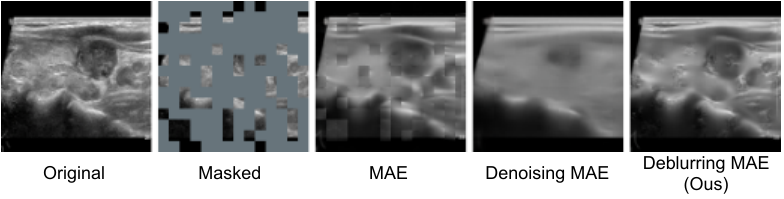}
  \vspace{-20pt}
  \caption{Comparisons of reconstructed images.}
  \label{fig_vis_compare}
\end{figure}

The comparisons of reconstructed image examples among MAE, denoising MAE, and our proposed deblurring MAE are illustrated in Figure~\ref{fig_vis_compare}. Although there is no strong evidence that reveals the relationship between reconstruction quality in pretraining and downstream task performance in MAE-based approaches, we can still obtain some insights from the reconstruction quality. As shown in Figure~\ref{fig_vis_compare}, we can clearly observe that the reconstructed images of the denoising MAE are the smoothest and lost most details among all the three approaches, followed by the vanilla MAE, and our deblurring MAE achieves the best reconstruction quality with much finer details. The comparisons indicate that our deblurring MAE can capture critical details that are beneficial for downstream classification. More comparisons can be found in the appendix.

\section{Conclusion and Future Work}
In this paper, we propose a novel deblurring MAE by incorporating deblurring into the proxy task during MAE pretraining for ultrasound image recognition. The deblurring task is implemented by inserting image blurring operation prior to the random masking during pretraining. The integration of deblurring enables the pretraining pay more attention to recovering the intricate details presented in ultrasound images, which are critical for downstream image classification. We explore the effect of several different image blurring methods and find that Gaussian blurring achieves the best performance and only a limited range of blurriness has a beneficial effect for pretraining. 
Based on the optimal blurring method and blurriness, our deblurring MAE achieves state-of-the-art performance in the downstream classification of ultrasound images, 
indicating the effectiveness of incorporating deblurring into MAE pretraining for ultrasound image recognition. 
However, this work has some limitations. For example, only one downstream task: nodule classification is evaluated in this study. We plan to extend our approach to include more tasks such as segmentation in the future.

\subsubsection{Acknowledgment.} This work was supported by Natural Science Foundation of Sichuan Province under Grant NO. 2022NSFSC1855.

%
%
%
\bibliographystyle{splncs04}

%




\end{document}


\appendix
\appendixpage

\section{Experimental Details}

\begin{table}[hbt!]
\centering
\caption{Pretraining setting.}
\label{table_set_pretrain}
\resizebox{0.65\textwidth}{!}{%
\begin{tabular}{l|l}
\hline \hline
config                 & value                       \\ \hline
optimizer              & AdamW                       \\
base learning rate     & 1.5e-4                      \\
weight decay           & 0.05                        \\
optimizer momentum     & $\beta_1$, $\beta_2$ = 0.9, 0.95 \\
batch size             & 256                         \\
learning rate schedule & cosine decay                \\
warmup epochs          & 40                          \\
augmentation           & RandomResizedCrop           \\
total training epochs  & 12000                       \\
\hline \hline
\end{tabular}%
}
\end{table}

\begin{table}[hbt!]
\centering
\caption{End-to-end fine-tuning setting.}
\label{table_set_finetune}
\resizebox{0.65\textwidth}{!}{%
\begin{tabular}{l|l}
\hline \hline
config                 & value                         \\ \hline
optimizer              & AdamW                         \\
base learning rate     & 1e-3                          \\
weight decay           & 0.05                          \\
optimizer momentum     & $\beta_1$, $\beta_2$ = 0.9, 0.999 \\
layer-wise lr decay    & 0.75                          \\
batch size             & 256                           \\
learning rate schedule & cosine decay                  \\
warmup epochs          & 5                             \\
augmentation           & RandAug (9, 0.5)              \\
label smoothing        & 0.1                           \\
mixup                  & 0.8                           \\
cutmix                 & 1.0                           \\
drop path              & 0.1 (B/L)  0.2 (H)            \\
 \hline \hline
\end{tabular}%
}
\end{table}

\begin{table}[hbt!]
\centering
\caption{Linear probing setting.}
\label{table_set_lineprobe}
\resizebox{0.65\textwidth}{!}{%
\begin{tabular}{l|l}
\hline \hline
config                 & value                       \\ \hline
optimizer              & LARS                        \\
base learning rate     & 0.1                         \\
weight decay           & 0                           \\
optimizer momentum     & 0.9                         \\
batch size             & 1024                         \\
learning rate schedule & cosine decay                \\
warmup epochs          & 10                          \\
augmentation           & RandomResizedCrop           \\
\hline \hline
\end{tabular}%
}
\end{table}

\section{Visualization}

\begin{figure}[hbt!]
  \centering
  \includegraphics[width=\textwidth]{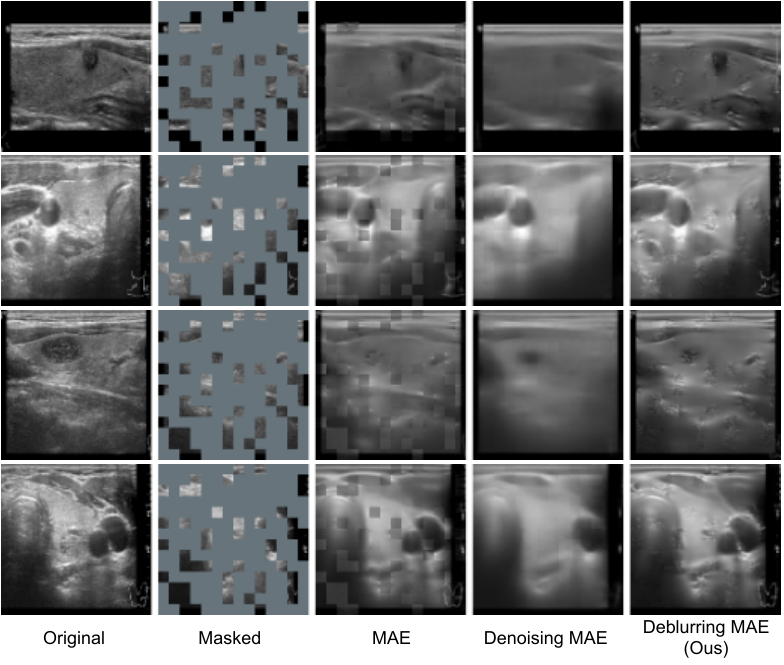}
  \caption{Comparisons of reconstruction among MAE, denoising MAE, and our proposed deblurring MAE.}
  \label{fig_reconst_vis}
\end{figure}